# Comparing Bayesian Network Classifiers

Jie Cheng   Russell Greiner
Department of Computing Science
University of Alberta
Edmonton, Alberta T6G 2H1 Canada
Email: {jcheng, greiner}@cs.ualberta.ca

## Abstract

In this paper, we empirically evaluate algorithms for learning four types of Bayesian network (BN) classifiers – Naïve-Bayes, tree augmented Naïve-Bayes, BN augmented Naïve-Bayes and general BNs, where the latter two are learned using two variants of a conditional-independence (CI) based BN-learning algorithm. Experimental results show the obtained classifiers, learned using the CI based algorithms, are competitive with (or superior to) the best known classifiers, based on both Bayesian networks and other formalisms; and that the computational time for learning and using these classifiers is relatively small. Moreover, these results also suggest a way to learn yet more effective classifiers; we demonstrate empirically that this new algorithm does work as expected. Collectively, these results argue that BN classifiers deserve more attention in machine learning and data mining communities.

## 1 INTRODUCTION

Many tasks – including fault diagnosis, pattern recognition and forecasting – can be viewed as *classification*, as each requires identifying the class labels for instances, each typically described by a set of features (attributes).

Learning accurate classifiers from pre-classified data is a very active research topic in machine learning and data mining. In the past two decades, many algorithms have been developed for learning decision-tree and neural-network classifiers. While Bayesian networks (BNs) (Pearl 1988) are powerful tools for knowledge representation and inference under conditions of uncertainty, they were not considered as classifiers until the discovery that Naïve-Bayes, a very simple kind of BNs that assumes the attributes are independent given the classification node, are surprisingly effective (Langley *et al.* 1992).

This paper further explores this role of BNs. Section 2 provides the framework of our research, describing standard approaches to learning simple Bayesian networks, then motivating our exploration for more effective extensions. Section 3 defines four classes of BNs – Naïve-Bayes, tree augmented Naïve-Bayes (TANs), BN augmented Naïve-Bayes (BANs) and general BNs (GBNs) – and describes methods for learning each. We implemented these learners to test their effectiveness. Section 4 presents and analyzes the experimental results, over a set of standard learning problems. We also propose a new algorithm for learning yet better BN classifiers, and present empirical results that support this claim.

## 2 FRAMEWORK

### 2.1 BAYESIAN NETWORKS

A Bayesian network $B = <N, A, \Theta>$ is a directed acyclic graph (DAG) $<N,A>$ with a conditional probability distribution (CP table) for each node, collectively represented by $\Theta$. Each node $n \in N$ represents a domain variable, and each arc $a \in A$ between nodes represents a probabilistic dependency (see Pearl 1988). In general, a BN can be used to compute the conditional probability of one node, given values assigned to the other nodes; hence, a BN can be used as a classifier that gives the *posterior probability distribution* of the classification node given the values of other attributes. When learning Bayesian networks from datasets, we use nodes to represent dataset attributes.

We will later use the idea of a *Markov boundary* of a node $n$ in a BN, where $n$'s Markov boundary is a subset of nodes that "shields" $n$ from being affected by any node outside the boundary. One of n's Markov boundaries is its *Markov blanket*, which is the union of $n$'s parents, $n$'s children, and the parents of $n$'s children. When using a BN classifier on complete data, the Markov blanket of the classification node forms a natural feature selection, as all features outside the Markov blanket can be safely deleted from the BN.

This can often produce a much smaller BN without compromising the classification accuracy.

## 2.2  LEARNING BN'S

The two major tasks in learning a BN are: learning the graphical structure, and then learning the parameters (CP table entries) for that structure. As it is trivial to learn the parameters for a given structure that are optimal for a given corpus of complete data – simply use the empirical conditional frequencies from the data (Cooper and Herskovits 1992) – we will focus on learning the BN structure.

There are two ways to view a BN, each suggesting a particular approach to learning. First, a BN is a structure that encodes the joint distribution of the attributes. This suggests that the best BN is the one that best fits the data, and leads to the *scoring-based* learning algorithms, that seek a structure that maximizes the Bayesian, MDL or Kullback-Leibler (KL) entropy scoring function (Heckerman 1995; Cooper and Herskovits 1992).

Second, the BN structure encodes a group of conditional independence relationships among the nodes, according to the concept of *d-separation* (Pearl 1988). This suggests learning the BN structure by identifying the conditional independence relationships among the nodes. Using some statistical tests (such as Chi-squared test and mutual information test), we can find the conditional independence relationships among the attributes and use these relationships as constraints to construct a BN. These algorithms are referred as *CI-based* algorithms or constraint-based algorithms (Spirtes and Glymour 1996; Cheng *et al.* 1997a).

Heckerman *et al.* (1997) compare these two general learning, and show that the scoring-based methods often have certain advantages over the CI-based methods, *in terms of modeling a distribution*. However, Friedman *et al.* (1997) show theoretically that the general scoring-based methods may result in poor *classifiers* since a good classifier maximizes a different function – *viz.*, classification accuracy. Greiner *et al.* (1997) reach the same conclusion, albeit via a different analysis. Moreover, the scoring-based methods are often less efficient in practice.

This paper further demonstrates that the CI-based learning algorithms do not suffer from these drawbacks when learning BN *classifiers*.

We will later use the famous Chow and Liu (1968) algorithm for learning *tree-like* BNs from complete data, which provably finds the optimal (in term of log-likelihood score) tree-structured network, using only $O(N^2)$ pair-wise dependency calculations – *n.b.*, without any heuristic search. Interestingly, this algorithm has the features of both scoring-based methods and the CI-based methods – although this algorithm does find a structure with the best score, it does this by analyzing the pair-wise dependencies. Unfortunately, for unrestricted BN learning, no such connection can be found between the scoring-based and CI-based methods.

## 2.3  SIMPLE BN CLASSIFIERS

### 2.3.1  Naïve-Bayes

A Naïve-Bayes BN, as discussed in (Duda and Hart, 1973), is a simple structure that has the classification node as the parent node of all other nodes (see Figure 1). No other connections are allowed in a Naïve-Bayes structure.

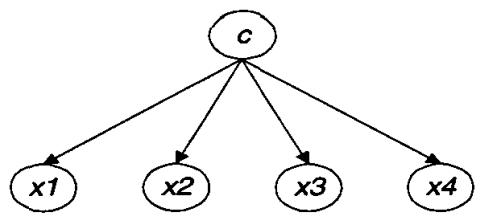

Figure 1: A simple Naïve Bayes structure

Naïve-Bayes has been used as an effective classifier for many years. It has two advantages over many other classifiers. First, it is easy to construct, as the structure is given *a priori* (and hence no *structure* learning procedure is required). Second, the classification process is very efficient. Both advantages are due to its assumption that all the features are independent of each other. Although this independence assumption is obviously problematic, Naïve-Bayes has surprisingly outperformed many sophisticated classifiers over a large number of datasets, especially where the features are not strongly correlated (Langley *et al.* 1992).

### 2.3.2  Improving on Naïve-Bayes

In recent years, a lot of effort has focussed on improving Naïve-Bayesian classifiers, following two general approaches: selecting feature subset and relaxing independence assumptions.

**Selecting Feature Subsets**

Langley and Sage (1994) use forward selection to find a good subset of attributes, then use this subset to construct a *selective Bayesian classifier* (ie, a Naïve-Bayesian classifier over only these variables). Kohavi and John (1997) use best-first search, based on accuracy estimates, to find a subset of attributes. Their algorithm can wrap around any classifiers, including either the decision tree classifiers or the Naïve-Bayesian classifier, etc. Pazzani's algorithm (Pazzani 1995) performs feature joining as well as feature selection to improve the Naïve-Bayesian classifier.

**Relaxing Independence Assumption**

Kononenko's algorithm (Kononenko 1991) partitions the attributes into disjoint groups and assumes independence only between attributes of different groups. Friedman et al. (1997) studied TAN, which allows tree-like structures to be used to represent dependencies among attributes (see Section 3.2). Their experiments show that TAN outperforms Naïve-Bayes, while maintaining computational simplicity on learning. Their paper also presents methods for learning GBNs and BANs (Sections 3.3, 3.4) as classifiers, using a minimum description length (MDL) scoring method when deciding among different structures.

**Combining the two approaches**

Singh and Provan (1996) combine the two approaches by first performing feature selection using information theoretic methods and then using a scoring based BN learning method to learn a *selective BN* from the subset of features.

### 2.4 MOTIVATIONS

Although each of these algorithms is often more accurate than Naïve Bayes classifiers, there is still a lot of work to be done. We are particularly interested in the following questions.

1. Since "*using MDL (or other nonspecialized scoring functions) for learning unrestricted Bayesian networks may result in poor classifier...*" (Friedman et al. 1997), a natural question is "Will *non-scoring* methods (i.e., condition independence (CI) test based methods, such as mutual information test and chi-squared test based methods) learn good classifiers?"

2. If we treat the classification node as an ordinary node and learn an unrestricted BN, we get a natural feature subset – the *Markov blanket* (Section 2.1) around the classification node. Can we take this approach to select features (rather than earlier approaches that perform feature selection before the BN-learning)?

3. GBNs and BANs have more parameters, which increases the risk of overfitting given relatively small training sets. How can we cope with this concern? (Overfitting is a phenomenon that occurs when a model tries to fit the training set too closely instead of generalizing. An overfitted model is one that does not perform well on data outside the training samples.)

4. Efficiency is a major reason for learning and using TAN or Naïve-Bayes. How expensive is it to learn and use GBN and BAN classifiers?

In this paper, we investigate these questions using an empirical study. We use two variants of a general BN-learning algorithm (based on conditional-independence tests) to learn GBNs and BANs. We empirically compared these classifiers with TAN and Naïve-Bayes using eight datasets from the UCI Machine Learning Repository (Murphy and Aha, 1995). Our results motivate a new type of classifier (a wrapper of the GBN and the BAN), which is also empirically evaluated.

## 3   LEARNING BAYESIAN NETWORK CLASSIFIERS

This section presents algorithms for learning four different (successively more general) types of Bayesian network classifiers, which differ based on the *structures* that are permitted. (Recall from Section 2.2 that learning the parameters, for a given structure, is straightforward.)

Most of these methods are implemented on top of the *PowerConstructor 2.0* program (Cheng 1998), which implements two BN-learning algorithms: one for the case when node ordering is given (the CBL1 algorithm – Cheng et al. 1997a); the other for the case when node ordering is not given (the CBL2 algorithm – Cheng et al. 1997b). (A node ordering specifies a total order of the nodes; we insist that no node can be an ancestor of a node that appears earlier in the order.)

In this paper, we treat the classification node as the first node in the ordering, and view the order of other nodes as arbitrary – we simply use the order they appear in the dataset. Therefore, we only need to use the CBL1 algorithm, which uses $O(N^2)$ mutual information tests ($N$ is the number of attributes in the dataset) and is linear in the number of cases. The efficiency is achieved by directly extending the Chow-Liu tree construction algorithm (Chow and Liu 1968) to a three-phase BN learning algorithm: *drafting*, which is essentially the Chow-Liu algorithm, *thickening*, which adds edges to the draft, and *thinning*, which verifies the necessity of each edge. Given a sufficient number of samples, this algorithm is guaranteed to learn the optimal structure, when the underlying model of the data is *DAG-Faithful* (Spirtes and Glymour 1996). For the correctness proof, complexity analysis and other detailed information, please refer to (Cheng, 1997a).

### 3.1   NAÏVE-BAYES

The procedure of learning Naïve-Bayes (Figure 1) is:

1. Let the classification node be the parent of all other nodes.

2. Learn the parameters (recall these are just the empirical frequency estimates) and output the Naïve-Bayes BN.

## 3.2 TREE AUGMENTED NAÏVE-BAYES (TAN)

Let $X = \{x_1,...x_n,c\}$ represent the node set (where $c$ is the classification node) of the data. The algorithm for learning TAN classifiers (Friedman et al. 1997) first learns a tree structure over $X \setminus \{c\}$, using mutual information tests conditioned on $c$. It then adds a link from the classification node to each feature node, similar to a Naïve-Bayes structure (i.e., the classification node is a parent of all other nodes) – see Figure 2. (Note that features $x_1,x_2,x_3,x_4$ form a tree.)

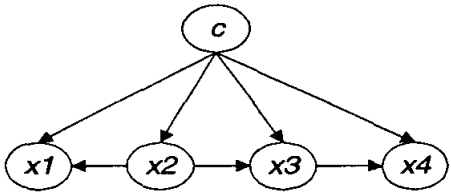

**Figure 2: A simple TAN structure**

The TAN learning procedure is:

1. Take the training set and $X \setminus \{c\}$ as input.

2. Call the modified Chow-Liu algorithm. (The original algorithm is modified by replacing every mutual information test $I(x_i,x_j)$ with a conditional mutual information test $I(x_i,x_j \mid \{c\})$.

3. Add $c$ as a parent of every $x_i$ where $1 \leq i \leq n$.

4. Learn the parameters and output the TAN.

This complete algorithm, which extends the Chow-Liu algorithm, requires $O(N^2)$ conditional mutual information tests.

## 3.3 BN AUGMENTED NAÏVE-BAYES (BAN)

BAN classifiers extend TAN classifiers by allowing the attributes to form an arbitrary graph, rather than just a tree (Friedman et al. 1997) -- see Figure 3. The BAN-learning algorithm is just like the TAN learner of Section 3.2, but the Step 2 of the BAN-learner calls an unrestricted BN-learning algorithm instead of a tree-learning algorithm.

Letting $X = \{x_1,...x_n,c\}$ represent the feature set (where $c$ is the classification node) of the data, the learning procedure based on mutual information test can be described as follows.

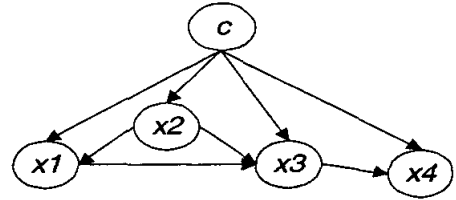

**Figure 3: A simple BAN structure**

1. Take the training set and $X \setminus \{c\}$ (along with the node ordering) as input.

2. Call a modified CBL1 algorithm – modified by replacing every mutual information test $I(x_i,x_j)$ with a conditional mutual information test $I(x_i,x_j \mid \{c\})$, and replacing every conditional mutual information test $I(x_i,x_j \mid Z)$ with $I(x_i,x_j \mid Z+\{c\})$, where $Z \subset X \setminus \{c\}$.

3. Add $c$ as a parent of every $x_i$ where $1 \leq i \leq n$.

4. Learn the parameters and output the BAN.

Like the TAN-learning algorithm, this algorithm does not require additional mutual information tests, and so it requires $O(N^2)$ mutual information tests.

## 3.4 GENERAL BAYESIAN NETWORK (GBN)

Unlike the other BN-classifier learners, the GBN learner treats the classification nodes as an ordinary node (see Figure 4). The learning procedure is described as follows.

1. Take the training set and the feature set (along with the node ordering) as input.

2. Call the (unmodified) CBL1 algorithm.

3. Find the *Markov blanket* of the classification node.

4. Delete all the nodes that are outside the Markov blanket.

5. Learn the parameters and output the GBN.

This algorithm also requires $O(N^2)$ mutual information tests.

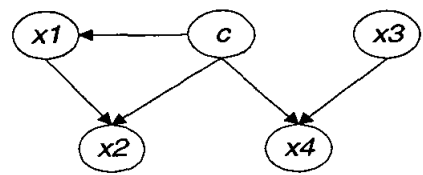

**Figure 4: A simple GBN**

# 4 EXPERIMENTS

## 4.1 METHODOLOGY

Our experiments were carried out using eight datasets downloaded from the UCI machine learning repository. When choosing the datasets, we selected datasets with large numbers of cases, to allow us to measure the learning and classification efficiency. We also preferred datasets that have few or no continuous features, to avoid information loss in discretization and to be able to compare the learning accuracy with other algorithms fairly. When we needed to discretize the continuous features, we used the discretization utility of MLC++ (Kohavi *et al.* 1994) on the default setting.

The datasets we used are summarized in Table 1. (CV5 stands for five-fold cross validation.)

**Table 1: Datasets used in the experiments.**

| Dataset | Attributes. | Classes | Instances | |
|---|---|---|---|---|
| | | | Train | Test |
| Adult | 13 | 2 | 32561 | 16281 |
| Nursery | 8 | 5 | 8640 | 4320 |
| Mushroom | 22 | 2 | 5416 | 2708 |
| Chess | 36 | 2 | 2130 | 1066 |
| DNA | 60 | 3 | 2000 | 1186 |
| Car | 6 | 4 | 1728 | CV5 |
| Flare | 10 | 3 | 1066 | CV5 |
| Vote | 16 | 2 | 435 | CV5 |

Brief descriptions of the eight datasets are given below.

*Adult dataset*: The data was extracted from the census bureau database. Prediction task is to determine whether a person makes over 50K a year. As the discretization process ignores one of the 14 attributes ("fnlwgt"), our learners therefore omitted "flnwgt" and used the rest 13 attributes in our experiments.

*Nursery*: Ranking nursery-school applications based on 8 features.

*Mushroom*: Classifying whether a type of mushroom is edible. Missing values are treated as having the value "?" in our experiments.

*Chess*: Chess end-game result classification based on board-descriptions.

*DNA*: Recognizing the boundaries between exons and introns given a sequence of DNA.

*Car dataset*: Car evaluation based on the six features of a car.

*Flare*: Classifying the number of times of occurrence of certain type of solar flare.

*Vote*: Using voting records to classify Congressmen as democrat and republican.

Our experiments were carried out as follows. We first used the four learning algorithms presented in Section 3 to learn the four classifiers (one of each type) and then export the BNs to *Bayesian Interchange Format* (BIF v0.15) files. The GBN and BAN classifiers were learned using the default threshold setting of the *PowerConstructor*. (This threshold determines how much mutual information between two nodes is considered as significant – see Section 4.3.) TAN and Naïve-Bayes learning algorithms have no such threshold.

To test these classifiers on the test sets, we use a modified version of *JavaBayes* (Cozman 1998). We added some classes to the *JavaBayes* v0.341 so that it can read the test datasets and perform classification given a BN.

The classification of each case in the test set is done by choosing, as class label, the value of class variable that has the highest posterior probability, given the instantiations of the feature nodes. The classification accuracy is measured by the percentage of correct predictions on the test sets (i.e., using a 0-1 loss function).

The experiments were performed using a Pentium II 300 MHz PC with 128MB of RAM, running MS-Windows NT 4.0.

## 4.2 RESULTS

Table 2 provides the prediction accuracy and standard deviation of each classifier. We ordered the datasets by their training sets from large to small. The best results of each dataset are emphasized using a boldfaced font. We also list the best results reported in the literature on these data sets (as far as we know).

From Table 2 we can see that the GBN, BAN and TAN have better overall performance than the Naïve-Bayes in these experiments. BAN did best on four of the datasets and GBN and TAN each did best on two of the datasets.

On the data sets "Nursery" and "Car", the GBN classifier was inferior to the Naïve-Bayes. The reason is, in both cases the GBN actually reduced to the Naïve-Bayes with missing links (the reduced Naïve-Bayes is sometimes called *selective Naïve-Bayes*). This reveals that the features in the two datasets are *almost* independent to each other. (Note that using *selective Naïve-Bayes* is not appropriate in such situations.) However, when using BAN or TAN classifiers, the weak dependencies among the features given the classification node can be successfully captured. These weak dependencies can improve the prediction accuracy significantly, as we can see from Table 2.

Since in GBN the classification node is treated in the same way as the feature nodes, these weak dependencies cannot be captured. This suggests that treating classification node differently is useful, at least in some domains.

We also measured the running time of the classifier-learning procedures. Table 3 gives the total learning time of each BN classifier, which includes the time for learning both structure and parameters. It shows that all BN classifiers can be learned efficiently as the longest learning time is less than 10 minutes. The unrestricted BN classifier learning procedures are at most five times slower than the TAN learning procedure. It took about 9 minutes to learn the BAN from the "Adult" data. Given that the training set has 32,561 cases and 13 attributes, the efficiency is quite satisfactory. Learning the BAN from the "DNA" data did take a long time (about 9.5 minutes), which suggests that the threshold here might be too small (see Section 4.3).

**Table 2: Running time (CPU seconds) of the classifier learning procedure.**

|          | GBN | BAN | TAN |
|----------|-----|-----|-----|
| Adult    | 515 | 536 | 131 |
| Nursery  | 10  | 11  | 10  |
| Mushroom | 136 | 134 | 38  |
| Chess    | 41  | 56  | 37  |
| DNA      | 300 | 570 | 113 |
| Car      | 1   | 1   | 1   |
| Flare    | 3   | 3   | 2   |
| Vote     | 8   | 8   | 3   |

In our experiments, we found that the classification process is also very efficient. *JavaBayes* can perform 100 to 600 classifications per second depending on the complexity of the classifier. We also found that the GBNs are often faster than Naïve-Bayes on classification when the GBNs contain only a subset of the features.

### 4.3 EXTENSIONS

As we mentioned earlier, the GBN and BAN classifiers were learned using the default threshold setting given by *PowerConstructor*. Based on our experience, this threshold setting is appropriate for most domains when the dataset is large enough. When the dataset is small, however, too high a setting will cause missing edges, and too low a setting will cause overfitting. Both situations will decrease the prediction accuracy of GBN and BAN. From Table 2, we can see that GBN or BAN can produce outstanding results when the datasets are large enough for their domains. For example, on the "Adult" data, the GBN gives the best result (as far as we know); on the "mushroom" data, the BAN gives 100% prediction accuracy. On the same dataset, the GBN gives 99.3% of accuracy using only 5 of the 22 features. However, when the datasets are small, there is evidence that show the threshold is not appropriate. For example, on the "Flare" data, the GBN uses only 1 to 3 of the 10 features, which is probably too harsh. As another example, note that the GBN and BAN learned from the "DNA" data perform worse than the TAN and Naïve classifiers. After examining the obtained GBN and BAN, we found that the structures are very complex, which suggests that the threshold is probably too small and the learner is overfitting. An advantage of the TAN classifier is that it does not need threshold. But we believed that, with a proper threshold setting, the unrestricted BN classifiers (GBN and BAN) should outperform the TAN classifier even when the dataset is small. In Section 4.4, we propose a wrapper algorithm that searches for the optimal threshold, and demonstrate its effectiveness on several of these datasets.

From Table 2, we can also find that, on six of the eight datasets, at least one of the unrestricted BN classifiers (i.e., GBN or BAN) give the best performance. This suggests that wrapping the two algorithms together and returning the winner will probably lead to better performance – again see Section 4.4.

### 4.4 AN IMPROVED BN CLASSIFIER

The analysis of our experimental results suggests two ways to improve the unrestricted BN classifiers:

- Automatic threshold selection based on the prediction accuracy.
- Wrapping the GBN and BAN together and returning the winner.

(Another algorithm for automatic threshold selection on BN construction is presented in [Fung and Crawford 1990].)

We therefore propose a wrapper algorithm that incorporates these two ideas.

1. Partition the input training set into internal training set and internal holdout set.

2. Call GBN-learner using different threshold settings and select a GBN that performs best on the holdout set.

3. Call BAN-learner using different threshold settings and select a BAN that performs best on the holdout set.

4. Select the better one of the two classifiers learned at Step 2 and Step 3.

Table 3: Experimental Results

|  | GBN (No. Selected Fea./Total No. Fea.) | BAN | TAN | Naïve-Bayes | Best reported |
|---|---|---|---|---|---|
| Adult | **86.11±0.27** (8/13) | 85.82±0.27 | 86.01±0.27 | 84.18±0.29 | 85.95 |
| Nursery | 89.72±0.46 (6/8) | **93.08±0.39** | 91.71±0.42 | 90.32±0.45 | N/A |
| Mushroom | 99.30±0.16 (5/22) | **100** | 99.82±0.08 | 95.79±0.39 | 100 |
| Chess | **94.65±0.69** (19/36) | 94.18±0.72 | 92.50±0.81 | 87.34±1.02 | 99.53±0.21 |
| DNA | 79.09±1.18 (43/60) | 88.28±0.93 | 93.59±0.71 | **94.27±0.68** | 96.12±0.6 |
| Car | 86.11±1.46 (5/6) | **94.04±0.44** | **94.10±0.48** | 86.58±1.78 | N/A |
| Flare | 82.27±1.45 (1-3/10) | 82.85±2.00. | **83.49±1.29** | 80.11±3.14 | 83.40±1.67 |
| Vote | 95.17±1.89 (10-11/16) | **95.63±3.85** | 94.25±3.63 | 89.89±5.29 | 96.3±1.3 |

5. Keep this classifier's structure and re-learn the parameters (conditional probability tables) using the whole training set.

6. Output this new classifier.

When the training set is not large enough, cross validation should be used to evaluate the performance of a classifier.

This wrapper algorithm is fairly efficient since it can reuse all the mutual information tests. Note that mutual information tests often take more than 95% of the running time of the BN learning process. The accuracy estimation process is not too slow either, since a few hundred predictions can be made in one second.

To test the performance of this wrapper, we use the largest five of the eight datasets. In each experiment, we use 2/3 of the training set as the internal training set and the rest 1/3 as the internal test set.

From Table 4 we can see that the wrapper is significantly better than the GBN and BAN on three of the five datasets. On the "DNA" dataset, by searching the optimal threshold, the prediction accuracy of GBN improved from about 80% to 96%, which makes this classifier one of the best results reported in the literature.

Another possible way of improving BN classifiers is by providing domain knowledge, such as node ordering, forbidden links and cause-&-effect relationships. This can help the BN learners to find a better BN structure. We might also use CBL2 to learn the BN classifiers without using prior node ordering, rather than specifying an arbitrary node ordering.

## 5 CONCLUSION

In this paper, we empirically evaluated and compared four BN classifiers. The unrestricted classifiers (GBN and BAN), learned using two variants of the CBL1 algorithm, give very encouraging results without using any wrapper function. After analyzing the experimental results, we proposed a wrapper algorithm that wraps around GBN and BAN, then demonstrated that this wrapper classifier can give even better results than the GBN and BAN classifiers.

Table 4 Experimental results using the wrapper

|  | GBN | | BAN | | Wrapper | Best reported |
|---|---|---|---|---|---|---|
|  | Fixed threshold | Optimal threshold | Fixed threshold | Optimal threshold | (best of the GBN and BAN using optimal thresholds) |  |
| Adult | **86.11±0.27** | unchanged | 85.82±0.27 | unchanged | **86.11±0.27** | 85.95 |
| Nursery | 89.72±0.46 | 90.32±0.45 | 93.08±0.39 | **95.74±0.31** | **95.74±0.31** | N/A |
| Mushroom | 99.30±0.16 | unchanged | **100** | unchanged | **100** | 100 |
| Chess | 94.65±0.69 | 94.09±0.72 | 94.18±0.72 | **96.44±0.57** | **96.44±0.57** | 99.53±0.21 |
| DNA | 79.09±1.18 | **95.95±0.57** | 88.28±0.93 | 94.69±0.65 | **95.95±0.57** | 96.12±0.6 |

From the experiments we can also see that the time expenses of unrestricted BN-learning are only (at most) a few times slower than that of the efficient TAN-learning. This is due to the three-phase learning mechanism used in the CBL1 algorithm.

Given the theoretical analysis (Friedman *et al.* 1997) and the empirical comparison (results using scoring-based methods on some of the data sets we use are reported in Friedman *et al.* 1997; Singh and Provan 1996), we believe that methods based on CI tests (such as mutual information tests) are more suitable for BN classifier learning than the more-standard scoring-based methods. Note, in addition, that such mutual information tests are standard in decision tree learning and feature selection.

Another advantage of these learners is that they are *constraint based*. This means we can incorporate domain knowledge relatively easily, by just adding them as additional constraints. Such additional information will lead to yet better classification accuracy.

Our experiments also show that the classification can be performed very efficiently using even-general BN classifiers.

Based on these results we believe that an improved type of BN classifiers, such as the ones shown here, should be used more often in real-world data mining and machine learning applications.